\pdfoutput=1%For ARXIV
\documentclass[lettersize,journal]{IEEEtran}
\usepackage{amsmath,amsfonts}
\usepackage{algorithmic}
\usepackage{algorithm}
\usepackage{array}
\usepackage[caption=false,font=normalsize,labelfont=sf,textfont=sf]{subfig}
\usepackage{textcomp}
\usepackage{stfloats}
\usepackage{url}
\usepackage{verbatim}
\usepackage{graphicx}
\usepackage{cite}
\usepackage{amssymb}
\usepackage{booktabs} 
\usepackage{multirow}
\usepackage{pifont}
\usepackage{color, colortbl}
\definecolor{Gray}{gray}{0.92}
\begin{document}

\title{Adaptive Base-class Suppression and Prior Guidance Network for One-Shot Object Detection}

\author{Wenwen~Zhang, Xinyu~Xiao, Hangguan~Shan and Eryun~Liu
        % <-this % stops a space
\thanks{This work was supported in part by the Zhejiang Provincial Natural Science Foundation of China under Grant LGF20F010006, and in part by the National Natural Science Foundation of China under Grant U21B2029 and Grant U21A20456.% <-this % stops a space

Wenwen Zhang, Hangguan Shan and Eryun Liu are with the College of Information Science and Electronic Engineering, Zhejiang University, Hangzhou 310027,
China (e-mail: wenwenzhang@zju.edu.cn; hshan@zju.edu.cn; eryunliu@zju.edu.cn). All correspondences should be directed to Eryun Liu.
Xinyu Xiao is currently a researcher with the Tencent company, Shanghai 200233, China (email: lawxiao@tencent.com).}}% <-this % stops a space}

\markboth{Journal of \LaTeX\ Class Files,~Vol.~14, No.~8, August~2021}%
{Shell \MakeLowercase{\textit{et al.}}: A Sample Article Using IEEEtran.cls for IEEE Journals}

\maketitle

\begin{abstract}
  One-shot object detection (OSOD) aims to detect all object instances towards the given category specified by a query image. Most existing studies in OSOD endeavor to explore effective cross-image correlation and alleviate the semantic feature misalignment, however, ignoring the phenomenon of the model bias towards the base classes and the generalization degradation on the novel classes. Observing this, we propose a novel framework, namely Base-class Suppression and Prior Guidance (BSPG) network to overcome the problem. Specifically, the objects of base categories can be explicitly detected by a base-class predictor and adaptively eliminated by our base-class suppression module. Moreover, a prior guidance module is designed to calculate the correlation of high-level features in a non-parametric manner, producing a class-agnostic prior map to provide the target features with rich semantic cues and guide the subsequent detection process. Equipped with the proposed two modules, we endow the model with a strong discriminative ability to distinguish the target objects from distractors belonging to the base classes. Extensive experiments show that our method outperforms the previous techniques by a large margin and achieves new state-of-the-art performance under various evaluation settings.
\end{abstract}


\section{Introduction}
\label{sec:intro}
\IEEEPARstart{B}{enefiting} from the flourishing of deep convolutional neural networks, object detection has made tremendous progress over the past few years\cite{NIPS2015_14bfa6bb,Redmon_2016_CVPR,He_2017_ICCV,liu2016ssd,9064498,9037090,8489974,9855518}. Nevertheless, most of the advanced methods heavily rely on large-scale labeled datasets\cite{deng2009imagenet,lin2014microsoft,everingham2010pascal}, and they may struggle with new applications in which novel-class objects are not witnessed during the training phase. Inspired by the human powerful cognitive ability to recognize new objects with only a few examples, few-shot learning (FSL) has emerged as a promising technique\cite{vinyals2016matching,ravi2016optimization,NIPS2017_cb8da676,sung2018learning,9097252,9749203,9654176,9419063,9638447}. FSL constructs models that can generalize to new classes with limited annotated data, providing a potential solution for object detection in scenarios with novel-class objects.

\begin{figure}[!t]
  \centering
  {\includegraphics[width=3.5in]{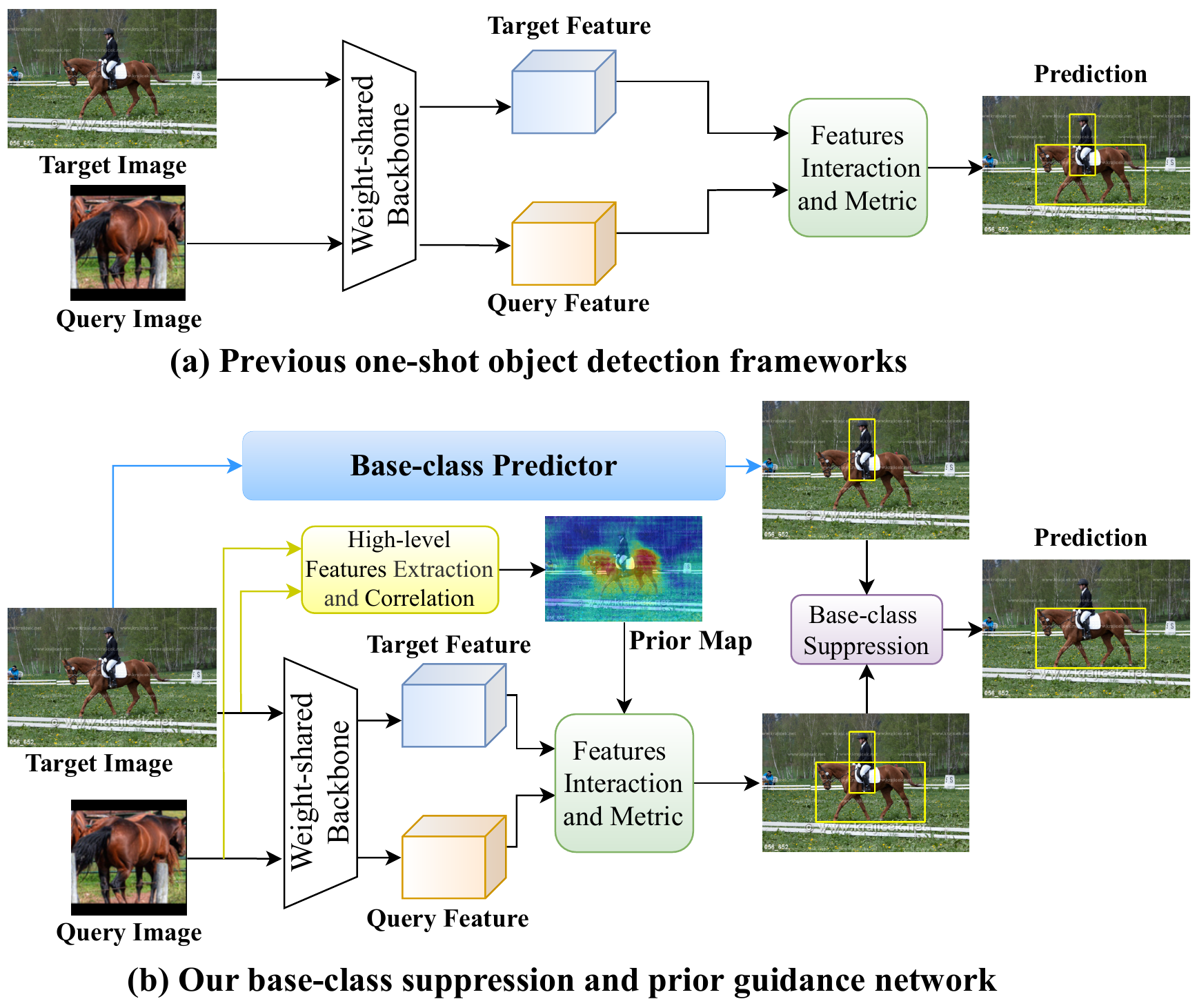}}
  \caption{Visualizations of the difference between our BSPG and previous OSOD works. (a) Existing OSOD models mostly exhibit a preference for familiar base-class objects (e.g., \textit{person}), leading to the performance degradation on the novel class specified by the query image patch (e.g., \textit{horse}). (b) We propose a base-class predictor to identify the base-class objects, a base-class suppression module to eliminate distractors, and a prior guidance module to produce a prior map to guide the detection process.}
  \label{fig:1}
  \end{figure}

  In this paper, we undertake the application of FSL in the field of object detection, termed one-shot object detection (OSOD), where the model aims to detect all instances of the given category specified by only one query image patch. In recent years, the field of few-shot object detection has thrived\cite{Yan_2019_ICCV,xiao2020few,Fan_2020_CVPR,Fan_2020_CVPR1,Zhu_2021_CVPR,Hu_2021_CVPR,9600874,9452164}, in which most prevalent approaches incorporate transfer-learning, meta-learning and metric-learning to deal with the task. The transfer-learning based model is first pre-trained on abundant base data and then fine-tuned on limited novel examples\cite{Chen_Wang_Wang_Qiao_2018,Sun_2021_CVPR,Fan_2021_CVPR}, which is time-consuming and application limited. The meta-learning based methods aim to learn meta knowledge for better generalization to novel classes. Most existing works in OSOD adopt the metric-learning frameworks and recognize new objects based on the similarity metrics between image-pairs without fine-tuning\cite{NEURIPS2019_92af93f7,Chen_2021_CVPR,Zhao_2022_CVPR}. However, they are generally dedicated to exploring effective cross-image feature interaction and mitigating semantic feature misalignment\cite{NEURIPS2019_92af93f7,Chen_2021_CVPR,9600874,Zhao_2022_CVPR}, neglecting the phenomenon of the model bias towards base classes and the generalization degradation on the novel classes\cite{Fan_2021_CVPR,Lang_2022_CVPR}. Due to the extremely unbalanced distribution of base-class and novel-class datasets, detectors tend to exhibit a preference for more familiar objects that lie in base classes rather than the given novel category. As illustrated in Fig.\,\ref{fig:1}(a), the conventional OSOD model easily suffers from the false positive detection on the objects of base categories, leading to the performance decrease on the novel categories.

In order to mitigate the above issues, we tackle the OSOD task from a new perspective. Instead of devising a more effective metric method, we improve the model by eliminating the distractors belonging to base classes. Specifically, a complementary branch named base-class predictor is introduced to explicitly detect the objects of base classes, which is pre-trained on the base dataset following a traditional paradigm. Furthermore, we observe that base-class predictor may produce some false predictions on the true novel objects due to interclass similarity. To avoid false suppression on the novel objects, we calculate the similarity between query and proposal features to guide the adjustment of the base-class results. Finally, we rectify the coarse detection results derived from the general OSOD network (novel-class predictor) by Base-class Suppression (BcS) module. As shown in Fig.\,\ref{fig:1}(b), equipped with the BcS module, the falsely detected objects are obviously suppressed, realizing accurate object detection for the specified new category.

As for the typical architecture of Faster R-CNN\cite{NIPS2015_14bfa6bb} in OSOD, the shallow and middle layers (\textit{res1-4}) of ResNet\cite{he2016deep} are regarded as a backbone to extract features for further interaction. While the deep layers (\textit{res5}) of ResNet are generally employed as an ROI feature extractor to encode proposal features. Inspired by the researches in the semantic segmentation field\cite{9154595,Zhao_2017_CVPR,Chen_2018_ECCV}, the high-level features convey rich semantic information and facilitate visual perception\cite{9154595}. In this work, ResNet-50 with deep layers (\textit{res1-5}) is applied as an additional backbone to encode the high-level features of the raw image-pair, where the backbone is pre-trained on the reduced ImageNet\cite{deng2009imagenet,NEURIPS2019_92af93f7,Chen_2021_CVPR} without further optimization. In the reduced ImageNet, we remove all COCO-related classes to guarantee that the model does not foresee the novel-class objects. Then we calculate the semantic relation between the high-level features to generate a class-agnostic prior map in a non-parametric manner, in which the regions of target features belonging to the co-existing objects can be activated, as illustrated in Fig.\,\ref{fig:1}(b). The prior map with rich semantic cues can provide guidance to promote the subsequent detection procedure and help the model distinguish target objects from backgrounds. Rather than pursuing sophisticated interaction, which may involve numerous operations and inevitably result in a bias towards the base classes, we fully exploit the high-level features and mine prior information to boost performance. As the prior generation is non-parametric, the model can learn more general patterns and retain generalization ability, thus implicitly alleviating the bias problem. With the two proposed modules described above, we establish a novel OSOD framework, termed Base-class Suppression and Prior Guidance (BSPG) network. These two components are complementary in terms of enhancing the generalization on novel classes.

Extensive experiments have been conducted on standard OSOD benchmarks to prove the effectiveness. Moreover, we extend it to few-shot settings, and our BSPG surpasses the advanced algorithms by a considerable margin. In summary, our contributions in this paper can be summarized as follows:
\begin{itemize}
\item We propose a novel and effective BSPG network to resolve the model bias towards base classes and enhance the generalization on novel classes, which has been neglected in previous work on OSOD.

\item We introduce a base-class predictor to identify the objects of base classes guided by a similarity map, and a base-class suppression module to adaptively eliminate them.

\item We propose a prior guidance module to generate a class-agnostic prior map via high-level features correlation, aiming to provide rich semantic cues and guide the subsequent detection procedure. 

\item Extensive experiments illustrate that our proposed framework yields new state-of-the-art results, which validate its effectiveness and robustness.

\end{itemize}

%------------------------------------------------------------------------
\section{Related Work}
\label{sec:Related}
\subsection{General Object Detection}
Object detection is a fundamental computer vision task that aims to locate objects of seen-classes and assign a category label to each object instance. General detectors can be broadly divided into two streams: one-stage methods and two-stage ones. The one-stage pipelines directly perform classification and bounding box regression based on a predefined set of anchor boxes\cite{Redmon_2017_CVPR,liu2016ssd,Lin_2017_ICCV} or anchor points\cite{Law_2018_ECCV,Duan_2019_ICCV,Tian_2019_ICCV,Zhang_2020_CVPR}. On the contrary, the two-stage proposal-based methods\cite{Girshick_2014_CVPR,Girshick_2015_ICCV,NIPS2015_14bfa6bb,He_2017_ICCV,Cai_2018_CVPR}, pioneered by R-CNN\cite{Girshick_2014_CVPR}, initially generate class-agnostic region proposals for coarsely locating the potential objects, and then classify them as well as refine their locations. Generally speaking, the two-stage detection methods have higher accuracy than the one-stage ones, while sacrificing detection efficiency. OSOD is still at its early stage of research, to pursue a high accuracy of algorithm, our model is developed based on the two-stage detector of Faster R-CNN\cite{NIPS2015_14bfa6bb}.

\subsection{Few-shot Learning}
Few-shot learning aims to learn a model that can be quickly applied to new tasks with only a small amount of labeled data. It has received extensive attention over the years due to its low cost of application. Generally, there are three mainstream methods, including the transfer-learning, optimization-based and metric-learning approaches. The transfer-learning approach\cite{qi2018low,gidaris2018dynamic} contains two stages. Initially, they learn backbone features by pretraining the model on base classes. Subsequently, they fine-tune the high-level layers using few new data to adapt to the novel classes. The optimization-based approach\cite{ravi2017optimization,finn2017model,lee2019meta} learns an initial model on the abundant training data and further quickly updates them with few data. The metric-learning method\cite{vinyals2016matching,NIPS2017_cb8da676,sung2018learning} utilizes a siamese network to extract feature embeddings of query and target images and computes their distance to predict whether the image-pairs belong to the same class. Despite tremendous progress in few-shot learning for classification, it remains challenging to adapt it to object detection, which involves not only class prediction but also localization. Our approach combines metric-learning with a general object detector to detect all object instances specified by the query image.

\subsection{Few-shot Object Detection}
Few-shot Object Detection (FSOD) performs object detection on a target image conditioned on a limited number of query images. Existing FSOD methods can be generally categorized into three directions: transfer learning, meta learning, and metric learning methods. For the transfer-learning based methods\cite{Chen_Wang_Wang_Qiao_2018,Sun_2021_CVPR,Qiao_2021_ICCV}, LSTD\cite{Chen_Wang_Wang_Qiao_2018} aims to utilize rich source-domain knowledge to develop a target-domain detector. FSCE\cite{Sun_2021_CVPR} integrates contrastive learning into Region-of-Interest (RoI) head to improve classification results. DeFRCN\cite{Qiao_2021_ICCV} performs decoupling among multiple modules of Faster R-CNN\cite{NIPS2015_14bfa6bb} to boost performance. The meta-learning based methods\cite{Yan_2019_ICCV,xiao2020few,9847356} are dedicated to learning efficient meta knowledge and fostering adaptation to novel categories. Meta R-CNN\cite{Yan_2019_ICCV} extends Faster R-CNN\cite{NIPS2015_14bfa6bb} by applying channel-wise attention to reweight the RoI features. FsDetView\cite{xiao2020few} builds on Meta R-CNN and designs a joint feature embedding module to improve detection results. The metric-learning based methods\cite{FU2021243,NEURIPS2019_92af93f7,ZHANG2022216,Chen_2021_CVPR,Zhao_2022_CVPR,9600874,Fan_2020_CVPR} focus on exploring cross-image correlations to directly detect novel objects without fine-tuning. Attention RPN\cite{Fan_2020_CVPR} introduces a multi-relation detector to measure the semantic relationship between query and proposals. DAnA\cite{9600874} proposes dual-awareness attention to better establish correlations. CoAE\cite{NEURIPS2019_92af93f7} adopts the non-local block\cite{Wang_2018_CVPR} and squeeze-excitation scheme\cite{Hu_2018_CVPR} to correlate the image-pairs. AIT\cite{Chen_2021_CVPR} excels in estimating the similarity benefiting from an attention-based encoder-decoder architecture. SaFT\cite{Zhao_2022_CVPR} deploys a unified attention mechanism to address semantic misalignment in space and scale. 

OSOD, as an extreme case of FSOD, involves the localization and classification for novel objects with only one sample. Recent researches \cite{Sun_2021_CVPR,YANG2021390} suggest that the box regressor is capable of accurately localizing novel instances. Owing to the seriously unbalanced datasets between base and novel classes, the main source of generalization degradation is misclassifying the instances of base classes as objects of interest. Therefore, we apply a base-class predictor to explicitly detect the base-class objects and further adaptively eliminate them, and perform prior guidance in a non-parametric manner to enhance the performance without sacrificing the generalization ability.

%-------------------------------------------------------------------------
\section{Method}
In this section, we first briefly introduce the problem definition of OSOD in Section~\ref{3_1}. Then we present the overall architecture of our model in Section~\ref{3_2}. Section~\ref{3_3} and \ref{3_4} respectively describe the proposed base-class suppression and prior guidance module in detail. 

\begin{figure*}
  \centering
  {\includegraphics[width=7.0in]{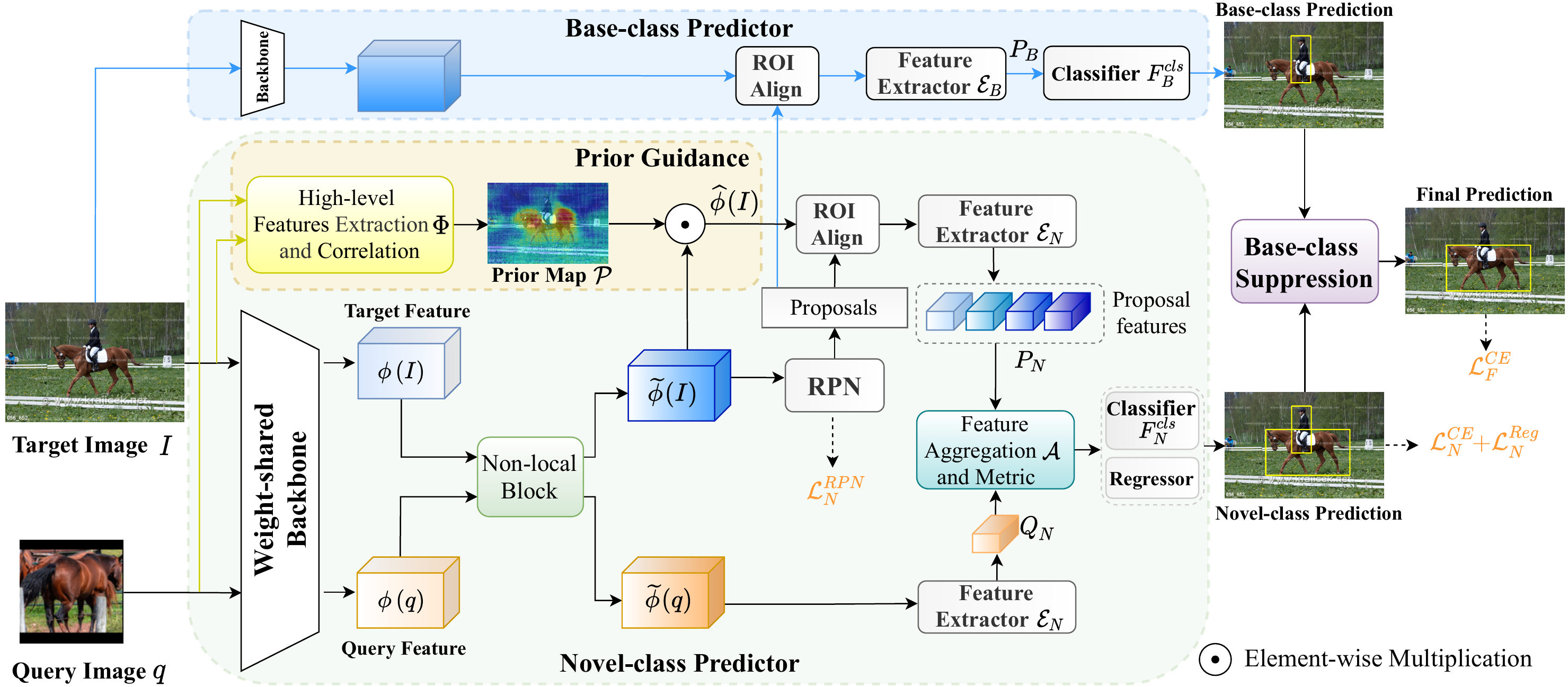}}
  \caption{The overall architecture of our proposed BSPG for one-shot object detection, which contains four key components: base-class predictor, novel-class predictor, base-class suppression module, and prior guidance module. Based on the base-class prediction, we refine the coarse novel-class prediction by adaptive base-class suppression to yield the final prediction results. Meanwhile, the prior guidance module generates a prior map as an indicator to provide rich semantic cues and guide the subsequent detection procedure. 
  }
    \label{fig:2}
  \hfill
\end{figure*}
%-------------------------------------------------------------------------
\subsection{Problem Definition}
\label{3_1}
Following the common configurations in previous literature\cite{NEURIPS2019_92af93f7,Chen_2021_CVPR,Zhao_2022_CVPR}, the object classes of the dataset are partitioned into two disjoint parts $\mathcal{C}_b\cap\mathcal{C}_n=\varnothing$, where $\mathcal{C}_b$ denotes base classes with abundant annotated data for training, and $\mathcal{C}_n$ represents novel classes with only one instance per category. We train our model with the episodic paradigm, where each episode is a target-query pair. The one-shot object detector is expected to detect all instances of the same category as specified by the query image $q$ in the target image $I$. The model is continuously optimized using numerous available data of base classes $\mathcal{C}_b$ during the training phase. Once the training is completed, the detector can directly predict the objects of novel classes $\mathcal{C}_n$ in the target image conditioned on one query image without fine-tuning.

%-------------------------------------------------------------------------
\subsection{Overview}
\label{3_2}
Fig.\,\ref{fig:2} sketches the overall architecture of our model, the base-class suppression and prior guidance (BSPG) network. It comprises four essential components, namely the base-class predictor, the novel-class predictor, the base-class suppression module, and the prior guidance module. We apply the two-stage training strategy to respectively train the base and novel predictors. In the first stage, we follow the traditional Faster R-CNN paradigm to optimize the base-class predictor on the base dataset. For the second stage, the backbone weights of the novel predictor are initialized from those in the base predictor, where only the last layer of the backbone (\textit{res4} in ResNet-50) is further optimized. We fix the parameters of the base predictor and optimize the novel predictor with base-class suppression. Specifically, the two predictors are deployed to respectively identify the objects of base and novel classes, where the base predictor uses the proposals generated by the novel predictor for convenience. Then, we rectify the coarse detection results derived from the novel predictor by adaptive base-class suppression to yield the final detection results. Meanwhile, we calculate the cross-image correlation between high-level features to produce the prior map, which serves as an indicator to provide discriminative semantic hints and guide the subsequent detection process. 

%-------------------------------------------------------------------------

%-------------------------------------------------------------------------
\subsection{Base-class Suppression}
\label{3_3}
\subsubsection{Base-class predictor}
To alleviate the performance degradation caused by mistaking base-class instances for target objects, we introduce an auxiliary branch called base-class predictor to explicitly detect the objects of base categories. However, the proposals generated by base-class predictor are mostly active in the base categories, which may burden the following classification task for novel-class detection. For simplicity, we only process the misclassification produced by the novel-class predictor. Therefore, the RPN of the base predictor is removed in the second training phase, and the candidate proposals generated by the novel predictor are delivered to the base predictor for further proposal feature extraction and classification. The classifier $F^{cls}_{B}$ of the base predictor is a fully connected layer, which yields the base-class classification results $Y_B$ for proposal features $P_B$:

\begin{equation}
Y_B = F^{cls}_{B}(P_B)\in \mathbb{R}{^{K \times (1+S)}},
  \label{eq1}
\end{equation}
where $K$ is the number of proposals (equaling 128 in our work), $S$ denotes the number of base categories (equaling 60 for COCO dataset\cite{lin2014microsoft}, and equaling 16 for VOC dataset\cite{everingham2010pascal} under one-shot settings), $Y_B$ denotes the prediction probability over $S$ categories for each proposal.

\subsubsection{Novel-class predictor}
Given a query image $q$ and a target image $I$, the novel-class predictor aims to detect the object instances in $I$ towards the given category specified by $q$. Specifically, the features $\phi(q)$, $\phi(I)$ extracted by the siamese ResNet-50 are aggregated by the non-local block\cite{Wang_2018_CVPR,NEURIPS2019_92af93f7}, intending to accomplish feature interaction. With more effective query information propagated to RPN, it can generate more expected proposals with high potential including target objects and filter out negative objects not belonging to the query category. Then we use the RoI Align layer\cite{He_2017_ICCV} and siamese feature extractor $\mathcal{E}_{N}$ (\textit{res5} in ResNet-50) to respectively obtain query feature $Q_N$ and the proposal features $P_N$ based on the candidate proposals.

To distinguish whether the proposals belong to the target category or not, we devise a feature aggregation network $\mathcal{A}$ consisting of vectors concatenation and feature maps correlation. The goal of this network is to measure the semantic relation and re-score proposals, formulated as Eq.\,(\ref{eq3}). Specifically, we concatenate the spatially global average pooling (GAP) vectors of each proposal feature and query feature. Complemented with the concatenation operation that performs global feature matching, the query feature $Q_N\in\mathbb{R}^{C \times H \times W}$ is regarded as a convolution kernel to generate the relevant features with each proposal feature $P_N\in\mathbb{R}^{C \times H \times W}$ in a depth-wise manner\cite{Li_2019_CVPR,Fan_2020_CVPR}, which performs pixel-wise feature matching. Then we deliver the aggregated features to the novel-class classifier $F^{cls}_{N}$ and obtain the coarse novel-class classification results $Y_N$.  

\begin{equation}
 \mathcal{A}(P_N,Q_N)=(GAP(P_N)\oplus GAP(Q_N), P_N\otimes Q_N),
  \label{eq3}
\end{equation}
\begin{equation}
Y_N = F^{cls}_{N}(\mathcal{A}(P_N,Q_N))\in \mathbb{R}{^{K \times 2}},
  \label{eq4}
\end{equation}
where $\oplus$ represents the concatenation operation along channel dimension, $\otimes$ represents the features correlation in a depth-wise manner.

\subsubsection{Base-class suppression}
Base-class suppression module is designed to explicitly eliminate the false predictions on the base classes. We choose the appropriate proposals whose highest confidence score over $S$ categories (excluding the background) are more than threshold $\alpha$ as the base-class objects $\mathcal{B}$:
\begin{equation}
\mathcal{B} = \max _{i \in\{1,2, \ldots, S\}}(Y_B^i)> \alpha,
  \label{eq5}
\end{equation}
where $\alpha$ is set to 0.7. Due to interclass similarity, the base-class predictor may occasionally produce false base-class positives but true novel-class positives, which subsequently leads to the false suppression on the novel-class objects. To address this, we calculate the cosine similarity between the proposal-query pairs, formulated as Eq.\,(\ref{eq6}), aiming to guide the adaptive adjustment of the raw base-class results.
\begin{equation}
 \mathcal{M}= \Theta(GAP(P_B))^T \Theta(GAP(Q_B)),
  \label{eq6}
\end{equation}
where $\mathcal{M}$ is the similarity map, $\Theta$ denotes the L2 normalization, $P_B$ and $Q_B$ respectively denote the proposal features and query feature encoded by the feature extractor $\mathcal{E}_{B}$ of the base predictor. A high similarity between proposal-query pair implies that the object is probably a true novel-class but false base-class positive. In contrast, the lower the similarity, the more likely it is indeed the distractor belonging to the base category. 

Finally, the coarse detection results $Y_N$ derived from the novel-class predictor are rectified by the base-class objects suppression, further yielding the final results $Y_F$:
\begin{equation}
Y_F = [\Psi_b(Y_N^b,\varphi(\mathcal{B},\mathcal{M}))\oplus \Psi_f(Y_N^f,\varphi(\mathcal{B},\mathcal{M}))]\in \mathbb{R}{^{K \times 2}},
  \label{eq7}
\end{equation}
where $Y_N^b$ and $Y_N^f$ respectively denote the background and foreground probability predicted by the novel predictor. $\Psi_b$, $\Psi_f$, and $\varphi$ are learnable fully connected layers, in which $\varphi$ is applied to adjust the raw base-class results guided by the similarity map, and $\Psi_b$, $\Psi_f$ are deployed to refine the coarse novel-class results by base-class suppression.

\begin{figure*}
  \centering
  {\includegraphics[width=7.0in]{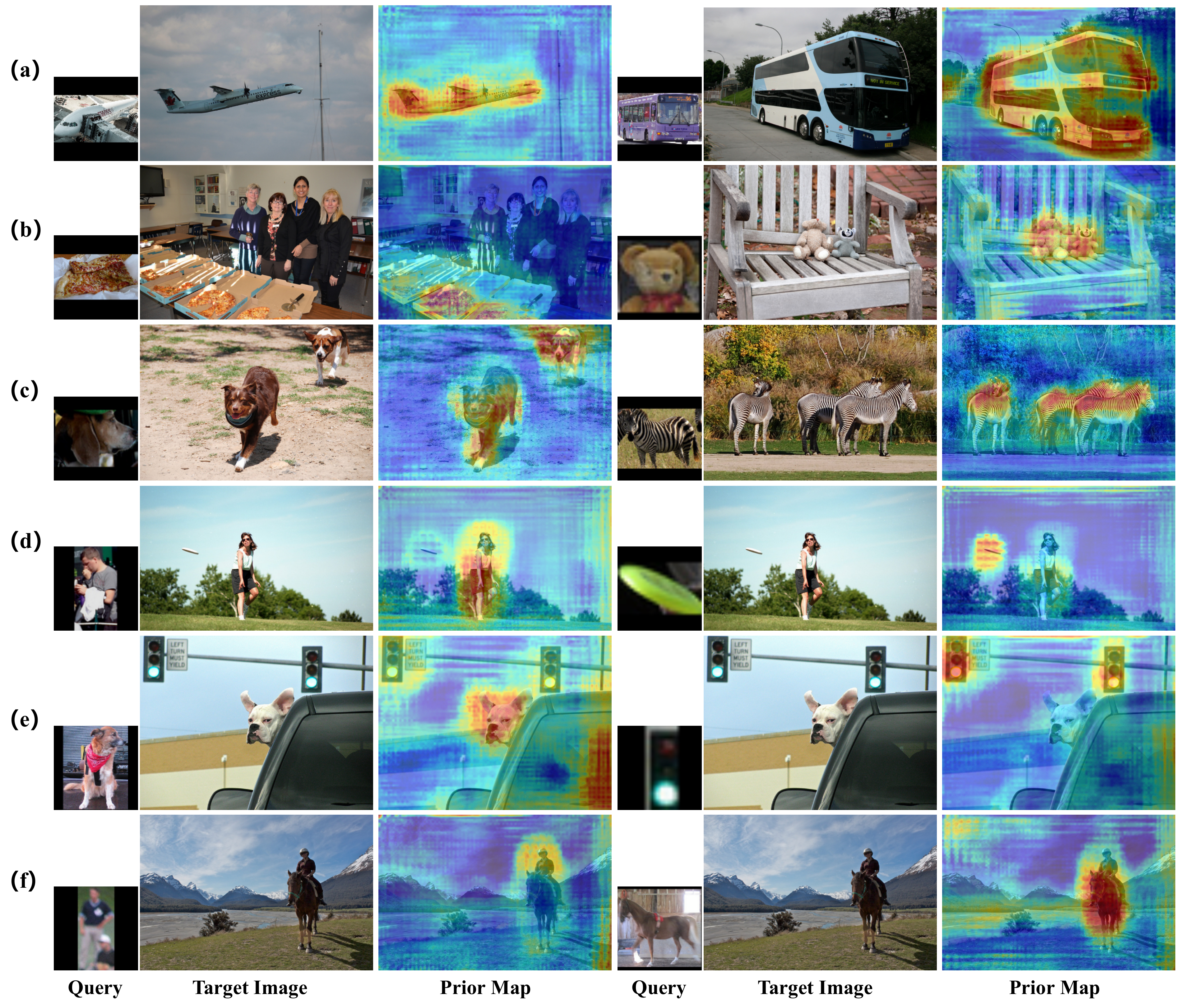}}
  \caption{Visualizations of the prior map generated by our prior guidance module via high-level features correlation. We show the query image, target image, and prior map, from left to right. The target images in groups (a-c) are different complex scenarios respectively including large objects, small objects, and multiple objects. For groups (d-f), each group has the same target image. Given different query images, the corresponding target regions in the prior maps can be respectively activated.}
  \label{fig:3}
  \hfill
\end{figure*}

\subsubsection{Loss function}
For the second stage, the overall loss for training our model can be expressed by:
\begin{equation}
  \mathcal{L}= \mathcal{L}^{RPN}_{N} + \mathcal{L}_{N}^{ROI},
   \label{eq12}
 \end{equation}
\begin{equation}
 \mathcal{L}_N^{ROI}=\lambda_1\mathcal{L}_{N}^{CE}(Y_N,G) + \lambda_2\mathcal{L}_{F}^{CE}(Y_F,G) +\mathcal{L}^{Reg}_{N},
  \label{eq8}
\end{equation}
where $\mathcal{L}^{RPN}_{N}$ is the RPN loss of Faster-RCNN\cite{NIPS2015_14bfa6bb}, and $\mathcal{L}_{N}^{ROI}$ is the loss in the ROI head of the novel predictor. $\mathcal{L}_{N}^{CE}$ and $\mathcal{L}_{F}^{CE}$ are cross entropy loss for evaluating the novel-class and final classification results, respectively. $G$ denotes the ground-truth label. $\lambda_1$ and $\lambda_2$ are empirically set to 0.5 in all experiments.

\subsection{Prior Guidance}
\label{3_4}
The backbone of the Faster R-CNN framework is composed of the shallow and middle layers (\textit{res1-4}) of ResNet, where the deep layers (\textit{res5}) are regarded as an ROI feature extractor to encode the proposal features for subsequent subtasks. Motivated by the researches in the semantic segmentation literature\cite{9154595,Zhao_2017_CVPR,Chen_2018_ECCV}, high-level features convey rich semantic information and benefit visual perception.  In view of this, our network employs the ResNet-50 including deep layers (\textit{res1-5}) as an additional backbone $\Phi$ to encode high-level features $\Phi(q)$ and $\Phi(I)$ of the raw image-pair, where the parameters are pre-trained on the reduced ImageNet\cite{deng2009imagenet,NEURIPS2019_92af93f7,Chen_2021_CVPR} without further optimization to retain generalization ability and avoid bias problem. Then we transform the high-level features into a class-agnostic prior map, which can reveal the semantic relation across the image-pair.

Specifically, we first calculate the element-wise relation map $\mathcal{R}\in\mathbb{R}^{H^h_I W^h_I\times H^h_q W^h_q}$ between $\Phi(I) \in\mathbb{R}^{C^h \times H^h_I \times W^h_I}$  and $\Phi(q)\in\mathbb{R}^{C^h \times H^h_q \times W^h_q} $ using cosine similarity function:
\begin{equation}
\label{eq9}
\mathcal{R} = \Theta(\Phi(I))^T \Theta(\Phi(q)).
\end{equation}

For each element in the target feature $\Phi(I)$, we select the maximum similarity among all elements of the query feature as the relation values to generate the prior map $\mathcal{P}\in\mathbb{R}^{H^h_I W^h_I\times 1}$, which can be summarized as: 
\begin{equation}
\label{eq10}
\mathcal{P} = {\max _{j \in\{1,2, \ldots, H^h_q W^h_q\}}\mathcal{R}(i,j)}.
\end{equation}

A high activation value in $\mathcal{P}$ for the element of the target feature indicates that this element has an intense correlation with at least one element in the query feature. As illustrated in Fig.\,\ref{fig:3}, the target images in groups (a-c) respectively include large objects, small objects, and multiple objects. Our prior map can serve as an indicator that coarsely locates the objects of interest despite the complex scenarios containing scale variations and appearance changes, which proves the effectiveness and robustness of our module. For groups (d-f), given different query images for the same target image, the corresponding target regions in the prior maps can be respectively activated, which validates the powerful generalization ability of our model. Notably, the above steps are independent of the training process to maintain generalization ability and prevent preference for base classes. 

Then, we normalize the values of the prior map to between 0 and 1, and reshape the map to match the shape of target feature $\widetilde{{\phi}}(I)$ output from the non-local block by an interpolation operation. Finally, the prior map is treated as guidance to reweight the target feature in the ROI head:
\begin{equation}
\label{eq11}
\widehat{\phi}(I)= \mathcal{P} \odot \widetilde{{\phi}}(I),
\end{equation}
where $\odot$ stands for element-wise multiplication.

\section{Experiments}
In this section, we conduct extensive experiments to evaluate our model. We begin by describing the details of the datasets and settings. Then, we compare our model with state-of-the-art methods following standard protocols. Ablation studies are also conducted to explore the impact of each component. Finally, we present qualitative results to provide an intuitive view of the effectiveness of the proposed approach.

\subsection{Datasets and Settings}
\subsubsection{Datasets and evaluation metrics}
For a fair comparison, we follow the protocol in previous works\cite{NEURIPS2019_92af93f7,Chen_2021_CVPR} to implement our method. Our model is tested on two widely used datasets, namely PASCAL VOC\cite{everingham2010pascal} and MS COCO\cite{lin2014microsoft}. For VOC dataset, the model is trained on VOC 2007 train$\&$val and VOC 2012 train$\&$val sets, and then is evaluated on VOC 2007 test set. The evaluation metric is the mean Average Precision (mAP) with IoU threshold 0.5 ($\text{AP}_{50}$). For COCO dataset, we train our model on COCO train 2017 set and evaluate on COCO val 2017 set with the standard COCO-style evaluation metrics that include AP, $\text{AP}_{50}$, and $\text{AP}_{75}$. The challenging COCO dataset has 80 object classes, consisting of a training set with 118,287 images and a validation set with 5000 images. 

\subsubsection{Implementation details}
Our training process consists of two phases, base-training and novel-training. Specifically, for the base-training phase, we adopt the ResNet-50 as our backbone and follow the general object detection paradigm to train the base-class predictor on each group of base classes. The backbone is pre-trained on the reduced ImageNet\cite{deng2009imagenet,NEURIPS2019_92af93f7,Chen_2021_CVPR}  where we remove all COCO-related classes to guarantee that the model does not foresee the novel-class objects. For the second phase, the parameters of the base predictor are fixed, and the backbone weights (\textit{res1-4}) of the novel predictor are initialized from the base predictor, where only the last layer (\textit{res4}) of the backbone is further optimized. For both phases, we optimize the network using the SGD optimizer with momentum of 0.9 for ten epochs, and with a batch size of 16 on two NVIDIA V100 GPUs in parallel. The learning rate starts at 0.01 and gradually decays by a ratio of 0.1 for every four epochs. 

\subsubsection{Target-query pairs}
We apply the same strategy as \cite{NEURIPS2019_92af93f7,Chen_2021_CVPR} to generate the target-query image pairs. In the novel-training stage, given a target image from the datasets, we randomly choose one query patch containing any of the same base-class in the target image. The episode of target-query pair forms the input to our model. In the testing stage, for each novel-class in a target image, the query images of the same class are shuffled with a random seed of target image ID, then the first five query images are respectively chosen to pair with the target image. We conduct evaluation on these image pairs and take the average of scores as the stable results. 

%-------------------------------------------------------------------------
\begin{table*}[]
  \setlength\tabcolsep{5pt}
  \caption{Performance comparison with OSOD methods on the COCO val dataset for novel classes in terms of AP, $\text{AP}_{50}$ and $\text{AP}_{75}$ (\%). $^\ast$ denotes its backbone is ResNet-101 with pre-trained weight from the original ImageNet. While other methods employ ResNet-50 as backbone and pre-train on the reduced ImageNet, where the novel classes are excluded.}
  \label{tab1}  
  \renewcommand\arraystretch{1} 
  \centering
  \begin{tabular}{c|ccccccccccccccc}
  \toprule
  \multirow{3}{*}{Method} & \multicolumn{15}{c}{Novel Classes} \\ \cline{2-16} 
   & \multicolumn{3}{c|}{split-1} & \multicolumn{3}{c|}{split-2} & \multicolumn{3}{c|}{split-3} & \multicolumn{3}{c|}{split-4} & \multicolumn{3}{c}{Average} \\ \cline{2-16} 
   & AP & $\text{AP}_{50}$ & \multicolumn{1}{c|}{$\text{AP}_{75}$} & AP & $\text{AP}_{50}$ & \multicolumn{1}{c|}{$\text{AP}_{75}$} & AP & $\text{AP}_{50}$ & \multicolumn{1}{c|}{$\text{AP}_{75}$} & AP & $\text{AP}_{50}$ & \multicolumn{1}{c|}{$\text{AP}_{75}$} & AP & $\text{AP}_{50}$ & $\text{AP}_{75}$ \\ \hline
  CoAE (NIPS2019)\cite{NEURIPS2019_92af93f7} & 11.8 & 23.4 & \multicolumn{1}{c|}{11.1} & 12.2 & 23.6 & \multicolumn{1}{c|}{11.7} & 9.3 & 20.5 & \multicolumn{1}{c|}{7.5} & 9.6 & 20.4 & \multicolumn{1}{c|}{8.2} & 10.7 & 22.0 & 9.6 \\
  AIT (CVPR2021)\cite{Chen_2021_CVPR} & - & 26.0 & \multicolumn{1}{c|}{-} & - & 26.4 & \multicolumn{1}{c|}{-} & - & 22.3 & \multicolumn{1}{c|}{-} & - & 22.6 & \multicolumn{1}{c|}{-} & - & 24.3 & - \\
  SaFT$^\ast$ (CVPR2022)\cite{Zhao_2022_CVPR} & - & \bf{27.8} & \multicolumn{1}{c|}{-} & - & 27.6 & \multicolumn{1}{c|}{-} & - & 21.0 & \multicolumn{1}{c|}{-} & - & 23.0 & \multicolumn{1}{c|}{-} & - & 24.9 & - \\
   BSPG (Ours) & \bf{14.4} & 25.8 & \multicolumn{1}{c|}{\bf{14.5}} & \bf{15.9} & \bf{28.2} & \multicolumn{1}{c|}{\bf{15.8}} & \bf{12.5} & \bf{23.8} & \multicolumn{1}{c|}{\bf{11.3}} & \bf{12.5} & \bf{23.4} & \multicolumn{1}{c|}{\bf{12.2}} & \bf{13.8} & \bf{25.3} & \bf{13.5} \\ \bottomrule
  \end{tabular}
  \end{table*}
  
  \begin{table*}[]
  \setlength\tabcolsep{5pt}
  \caption{Performance comparison with OSOD methods on the COCO val dataset for base classes in terms of AP, $\text{AP}_{50}$ and $\text{AP}_{75}$ (\%).}
  \label{tab2}  
  \renewcommand\arraystretch{1} 
  \centering
  \begin{tabular}{c|ccccccccccccccc}
  \toprule
  \multirow{3}{*}{Method} & \multicolumn{15}{c}{Base Classes} \\ \cline{2-16} 
   & \multicolumn{3}{c|}{split-1} & \multicolumn{3}{c|}{split-2} & \multicolumn{3}{c|}{split-3} & \multicolumn{3}{c|}{split-4} & \multicolumn{3}{c}{Average} \\ \cline{2-16} 
   & AP & $\text{AP}_{50}$ & \multicolumn{1}{c|}{$\text{AP}_{75}$} & AP & $\text{AP}_{50}$ & \multicolumn{1}{c|}{$\text{AP}_{75}$} & AP & $\text{AP}_{50}$ & \multicolumn{1}{c|}{$\text{AP}_{75}$} & AP & $\text{AP}_{50}$ & \multicolumn{1}{c|}{$\text{AP}_{75}$} & AP & $\text{AP}_{50}$ & $\text{AP}_{75}$ \\ \hline
  CoAE (NIPS2019)\cite{NEURIPS2019_92af93f7} & 22.4 & 42.2 & \multicolumn{1}{c|}{21.6} & 21.3 & 40.2 & \multicolumn{1}{c|}{20.6} & 21.6 & 39.9 & \multicolumn{1}{c|}{21.3} & 22.0 & 41.3 & \multicolumn{1}{c|}{21.3} & 21.8 & 40.9 & 21.2 \\
  AIT (CVPR2021)\cite{Chen_2021_CVPR} & - & 50.1 & \multicolumn{1}{c|}{-} & - & 47.2 & \multicolumn{1}{c|}{-} & - & 45.8 & \multicolumn{1}{c|}{-} & - & 46.9 & \multicolumn{1}{c|}{-} & - & 47.5 & - \\
  SaFT (CVPR2022)\cite{Zhao_2022_CVPR} & - & 49.2 & \multicolumn{1}{c|}{-} & - & 47.2 & \multicolumn{1}{c|}{-} & - & 47.9 & \multicolumn{1}{c|}{-} & - & 49.0 & \multicolumn{1}{c|}{-} & - & 48.3 & - \\
   BSPG (Ours) & \bf{30.8} & \bf{52.2} & \multicolumn{1}{c|}{\bf{31.9}} & \bf{29.7} & \bf{50.0} & \multicolumn{1}{c|}{\bf{30.7}} & \bf{30.7} & \bf{50.8} & \multicolumn{1}{c|}{\bf{32.3}} & \bf{30.7} & \bf{51.3} & \multicolumn{1}{c|}{\bf{31.9}} & \bf{30.5} & \bf{51.1} & \bf{31.7} \\ \bottomrule
  \end{tabular}
  \end{table*}

  \begin{table*}
    \setlength\tabcolsep{2.5pt}
    \caption{Performance comparison with OSOD methods on the PASCAL VOC dataset in terms of $\text{AP}_{50}$ score (\%).}
    \centering
    \label{tab2_voc}
    \begin{tabular}{ccccccccccccccccccccccc}
    \toprule
    \multirow{2}{*}{Method} & \multicolumn{17}{c|}{Base Class} & \multicolumn{5}{c}{Novel Class} \\ \cline{2-23} 
     & plant & sofa & tv & car & bottle & boat & chair & person & bus & train & horse & bike & dog & bird & mbike & \multicolumn{1}{c|}{table} & \multicolumn{1}{c|}{mAP} & cow & sheep & cat & \multicolumn{1}{c|}{aero} & mAP \\ \hline 
    SiamFC\cite{cen2018fully} &3.2	&22.8&	5.0&	16.7&	0.5&	8.1	&1.2&	4.2	&22.2&	22.6&	35.4	&14.2&	25.8&	11.7&	19.7&	\multicolumn{1}{c|}{27.8} &\multicolumn{1}{c|}{15.1} & 6.8&	2.28&	31.6&	\multicolumn{1}{c|}{12.4}&	13.3 \\
    SiamRPN\cite{li2018high} & 1.9&	15.7&	4.5&	12.8&	1.0&	1.1&	6.1&	8.7&	7.9&	6.9&	17.4&	17.8&	20.5&	7.2	&18.5&	\multicolumn{1}{c|}{5.1} &\multicolumn{1}{c|}{9.6} & 15.9&	15.7&	21.7&	\multicolumn{1}{c|}{3.5}	&14.2 \\
    OSCD\cite{fu2021oscd} & 28.4&	41.5&	65.0&	66.4&	37.1&	49.8&	16.2&	31.7&	69.7&	73.1&	75.6&	71.6&	61.4&	52.3&	63.4&	\multicolumn{1}{c|}{39.8}&\multicolumn{1}{c|}{52.7} & 75.3&	60.0&	47.9&	\multicolumn{1}{c|}{25.3}&	52.1\\
    CoAE\cite{NEURIPS2019_92af93f7} & 24.9&	50.1&	58.8&	64.3&	32.9&	48.9&	14.2&	53.2&	71.5&	74.7&	74.0&	66.3&	75.7&	61.5&	68.5&	\multicolumn{1}{c|}{42.7}&\multicolumn{1}{c|}{55.1} & 78.0&	61.9&	72.0&	\multicolumn{1}{c|}{43.5}&	63.8 \\
    AIT~\cite{Chen_2021_CVPR} & 46.4 &	60.5 &	68.0 & 73.6 &	49.0&	65.1&	26.6&	68.2&	82.6&	\bf85.4 &	82.9&	77.1 &	\bf82.7 &	71.8 &	75.1 & 
    \multicolumn{1}{c|}{60.0}&\multicolumn{1}{c|}{67.2} & 85.5 & \bf72.8 &	80.4&	\multicolumn{1}{c|}{\bf50.2}& \bf72.2\\
    BSPG (Ours) & \bf50.1 & \bf69.2 & \bf75.6 & \bf77.9 & \bf53.8 & \bf67.4 & \bf33.8 & \bf71.5 & \bf84.2 & 85.0 & \bf84.0 & \bf80.0 & 82.5 & \bf72.7 & \bf75.7 &\multicolumn{1}{c|}{\bf63.2}& \multicolumn{1}{c|}{\bf70.4} & \bf86.2 & 71.8 & \bf81.2 & \multicolumn{1}{c|}{47.9}& 71.8 \\ \bottomrule
    \end{tabular}
    \end{table*}

\begin{table*}[]
  \setlength\tabcolsep{7pt}
  \caption{Performance comparison with FSOD methods on the COCO val dataset for novel classes in terms of AP, $\text{AP}_{50}$ and $\text{AP}_{75}$ (\%). $^\dag$ represents the results reported in DAnA\cite{9600874}.}
  \label{tab3}
  \renewcommand\arraystretch{1} 
  \centering   
  \begin{tabular}{c|c|ccc|ccc|ccc}
  \toprule
  \multirow{2}{*}{Method} & \multirow{2}{*}{Category of classification} & \multicolumn{3}{c|}{1-shot} & \multicolumn{3}{c|}{3-shot} & \multicolumn{3}{c}{5-shot} \\ \cline{3-11} 
   &  & AP & $\text{AP}_{50}$ & $\text{AP}_{75}$ & AP & $\text{AP}_{50}$ & $\text{AP}_{75}$ & AP & $\text{AP}_{50}$ & $\text{AP}_{75}$ \\ \hline
   FSDetView (ECCV2020)\cite{xiao2020few} & Multi-classification & 4.5 & - & - & 7.2 & - & - & 10.7 & 24.5 & 6.7 \\
   Meta-DETR (TPAMI2022)\cite{9847356} & Multi-classification & 7.5 & 12.5 & 7.7 & 13.5 & 21.7 & 14.0 & 15.4 & 25.0 & 15.8 \\
   DeFRCN (ICCV2021)\cite{Qiao_2021_ICCV} & Multi-classification & 9.3 & - & - & 14.8 & - & - & 16.1 & - & - \\ \hline
   FGN$^\dag$ (CVPR2020)\cite{Fan_2020_CVPR1} & Two-classification & 8.0 & 17.3 & 6.9 & 10.5 & 22.5 & 8.8 & 10.9 & 24.0 & 9.0 \\
  Attention RPN$^\dag$ (CVPR2020)\cite{Fan_2020_CVPR} & Two-classification & 8.7 & 19.8 & 7.0 & 10.1 & 23.0 & 8.2 & 10.6 & 24.4 & 8.3 \\
  DAnA (TMM2021)\cite{9600874} & Two-classification & 11.9 & 25.6 & 10.4 & 14.0 & 28.9 & 12.3 & 14.4 & 30.4 & 13.0 \\
  BSPG (Ours) & Two-classification & \bf{13.5} & \bf{26.9} & \bf{12.4} & \bf{15.9} & \bf{30.8} & \bf{15.1} & \bf{17.4} & \bf{33.5} & \bf{16.4} \\ \bottomrule
  \end{tabular}
  \end{table*}      

\subsection{Comparison with State-of-the-art Methods}
Extensive comparisons with previous methods under different settings are performed to manifest the effectiveness and robustness of our method.

\subsubsection{Comparison with OSOD methods}
Our approach is mainly oriented towards complex scenarios consisting of multiple categories. The challenging COCO dataset whose images generally contain diverse objects is suitable for validating our ideas. Following the common practice as existing OSOD techniques\cite{NEURIPS2019_92af93f7,Chen_2021_CVPR,Zhao_2022_CVPR}, we equally divide the 80 classes into four groups, and take three groups as base classes and one group as novel classes in turns. As shown in Tabs.\,\ref{tab1} and \ref{tab2}, our model substantially outperforms the advanced methods by a considerable margin in most cases and reaches new state-of-the-art performance for both novel classes and base classes. The gains primarily derive from the fact that our model can effectively suppress the distractors belonging to the base classes and address the bias phenomenon. 

For VOC dataset, consistent with existing methods\cite{cen2018fully,li2018high,Zhao_2022_CVPR,fu2021oscd,NEURIPS2019_92af93f7,Chen_2021_CVPR}, the 20 classes are divided into 16 base and 4 novel classes, where the novel classes are \{‘‘aero”, ‘‘cat”, ‘‘cow”, ‘‘sheep”\}. From the results presented in Tabs.\,\ref{tab2_voc} we observe that the performance of our method on novel classes is comparable with the previous AIT model. We argue that our main contribution is to explicitly detect and suppress the distractors to boost performance, which is especially effective in scenes consisting of both base and novel classes. However, most images in VOC exhibit relatively simple scenarios with no distractors from base classes to be suppressed, leading to insignificant improvements.

\subsubsection{Comparison with FSOD methods}
Additionally, our model can be easily extended to few-shot settings. When several query images are available, we simply take the average of multiple query features before interacting with target features. We compare our BSPG with other advanced FSOD methods on two datasets and strictly adopt the same protocol to guarantee a fair comparison. Note that some FSOD algorithms\cite{9847356,xiao2020few,Qiao_2021_ICCV} consider the task as a multi-classification and localization problem. They first train the model on abundant base data and then fine-tune the model on limited novel data. Unlike these fine-tuning approaches, we treat the task as a two-classification and localization problem to better simulate the real-world application\cite{Fan_2020_CVPR,9600874}, which aims to retrieve novel objects towards reference category guided by a query image without fine-tuning. For COCO dataset, the experiments are implemented with the same data split in \cite{9847356,9600874,xiao2020few,Qiao_2021_ICCV}, where the 20 categories overlapped with PASCAL VOC\cite{everingham2010pascal} are treated as novel classes. As shown in Tab.\,\ref{tab3}, the proposed approach outperforms previous models among all settings, and surpasses the previous best competitor DAnA\cite{9600874} by 1.3$\%$, 1.9$\%$, 3.1$\%$ $\text{AP}_{50}$ under 1-shot, 3-shot, and 5-shot settings, respectively, verifying the superiority of our approach. For VOC dataset, we follow the three different split settings used as in \cite{xiao2020few,9847356,Qiao_2021_ICCV}, where 15 categories are treated as base classes, and the left 5 categories as novel classes. The results are reported in Tab.\,\ref{tab1_voc}, the proposed BSPG outperforms other approaches in most cases, evidently demonstrating that our method can be generalized to various conditions and achieve superior performance.

The consistent improvements in both OSOD and FSOD fields indicate that our model possesses a remarkable detection capability to identify new objects, which benefited from the effective base-class suppression and prior map guidance. Moreover, resolving the model bias towards base classes is a promising direction worth exploring.

\subsection{Ablation Study}
We conduct a series of ablation studies to investigate the impact of each key component. Following the previous work\cite{NEURIPS2019_92af93f7}, all related experiments are conducted on the split-2 of COCO dataset for novel classes under the 1-shot setting.

\begin{table*}[]
  \setlength\tabcolsep{6pt}
  \caption{Performance comparison with FSOD methods on the PASCAL VOC dataset for novel classes in terms of $\text{AP}_{50}$ score (\%). $^\ddagger$ represents the results reported in hANMCL\cite{park2022hierarchical}.}
  \label{tab1_voc}  
  \renewcommand\arraystretch{1} 
  \centering  
  \begin{tabular}{c|c|ccc|ccc|ccc}
  \toprule
  \multirow{2}{*}{Method / Shot} & \multirow{2}{*}{Category of classification} & \multicolumn{3}{c|}{Novel Set 1} & \multicolumn{3}{c|}{Novel Set 2} & \multicolumn{3}{c}{Novel Set 3} \\ \cline{3-11} 
   &  & 1 & 3 & 5 & 1 & 3 & 5 & 1 & 3 & 5 \\ \hline
   FSDetView (ECCV2020)\cite{xiao2020few} & Multi-classification & 24.2 & 42.2 & 49.1 & 21.6 & 31.9 & 37.0 & 21.2 & 37.2 & 43.8 \\
   DCNet (CVPR2021)\cite{Hu_2021_CVPR} & Multi-classification & 33.9 & 43.7 & 51.1 & 23.2 & 30.6 & 36.7 & 32.3 & 39.7 & 42.6 \\
   FSCE (CVPR2021)\cite{Sun_2021_CVPR} & Multi-classification & 32.9 & 46.8 & 52.9 & 23.7 & 38.4 & 43.0 & 22.6 & 39.5 & 47.3 \\
   Meta-DETR (TPAMI2022)\cite{9847356} & Multi-classification & 35.1 & 53.2 & 57.4 & 27.9 & 38.4 & 43.2 & 34.9 & 47.1 & 54.1 \\
   DeFRCN (ICCV2021)\cite{Qiao_2021_ICCV} & Multi-classification & 53.6 & 61.5 & 64.1 & 30.1 & 47.0 & \bf{53.3} & 48.4 & 52.3 & 54.9 \\ \hline
  FGN$^\ddagger$ (CVPR2020)\cite{Fan_2020_CVPR1} & Two-classification & 46.6 & 54.3 & 57.7 & 34.6 & 40.7 & 43.5 & 42.5 & 46.9 & 47.8 \\
  Attention RPN$^\ddagger$ (CVPR2020)\cite{Fan_2020_CVPR} & Two-classification & 48.1 & 54.6 & 56.4 & 33.1 & 41.0 & 41.5 & 43.0 & 48.5 & 49.1 \\
  DAnA$^\ddagger$ (TMM2021)\cite{9600874} & Two-classification & 51.7 & 57.8 & 60.1 & 33.4 & 41.4 & 43.0 & 40.8 & 48.3 & 53.4 \\
  hANMCL (Arxiv2022) \cite{park2022hierarchical} & Two-classification & 53.6 & 59.7 & 62.3 & 35.7 & 45.1 & 46.8 & 42.8 & 52.9 & 56.6 \\
  BSPG (Ours) & Two-classification & \bf{58.9} & \bf{63.3} & \bf{64.5} & \bf{39.8} & \bf{47.1} & 49.4 & \bf{53.3} & \bf{55.4} & \bf{59.5} \\ \bottomrule
  \end{tabular}
  \end{table*}
  
  \begin{table*}[]
    \setlength\tabcolsep{5pt}
    \caption{The number of FPs with respect to base classes and precision before and after BcS module, where the IoU denotes the intersection over union between the final predicted bounding box and the ground truth, the score denotes the corresponding predicted score. The experiments are performed on the split-2 of COCO val dataset, which contains 3309 test images.}
    \label{tab4_process1}  
    \renewcommand\arraystretch{1} 
    \centering     
    \begin{tabular}{c|ccl|ccl|cc|cc}
    \toprule
    \multirow{2}{*}{Method} & \multicolumn{3}{c|}{IoU\textgreater{}0.5, score\textgreater{}0.5} & \multicolumn{3}{c|}{IoU\textgreater{}0.5, score\textgreater{}0.75} & \multicolumn{2}{c|}{IoU\textgreater{}0.75, score\textgreater{}0.5} & \multicolumn{2}{c}{IoU\textgreater{}0.75, score\textgreater{}0.75} \\ \cline{2-11} 
     & FPs & \multicolumn{2}{c|}{Precision (\%)} & FPs & \multicolumn{2}{c|}{Precision (\%)} & FPs & Precision (\%) & FPs & Precision (\%) \\ \hline
    Before BcS module & 2174 & \multicolumn{2}{c|}{47.9} & 900 & \multicolumn{2}{c|}{56.7} & 1392 & 49.2 & 660 & 57.2 \\
    After BcS module & 1118 & \multicolumn{2}{c|}{62.5} & 593 & \multicolumn{2}{c|}{70.5} & 536 & 69.5 & 321 & 75.3 \\ \bottomrule
    \end{tabular}
    \end{table*}
    
\begin{table}[]
  \setlength\tabcolsep{12pt}
  \caption{Ablation study for each module in our model.}
  \label{tab4}  
  \renewcommand\arraystretch{1} 
  \centering
  \begin{tabular}{ll|lll}
  \toprule
  BcS & PG & AP & $\text{AP}_{50}$ & $\text{AP}_{75}$ \\ \hline
   &  & 14.5 & 25.5 & 14.9 \\
  \ding{51} &  & 15.5 & 27.4 & 15.4 \\
   & \ding{51} & 15.0 & 26.6 & 15.1 \\
    \ding{51} & \ding{51} & \bf{15.9} & \bf{28.2} & \bf{15.8} \\ \bottomrule
  \end{tabular}
  \end{table}
  
  \begin{table}[]
    \setlength\tabcolsep{6pt}
    \caption{Ablation study for different strategies in the proposed modules. ``W/ adjustment" denotes the base-class prediction adjustment guided by a similarity map. ``Cat-RPN/Mult-RPN/Cat-ROI/Mult-ROI" denote prior guidance via concatenation or multiplication in the RPN or ROI head.}
    \label{tab5}    
    \renewcommand\arraystretch{1} 
    \centering  
    \begin{tabular}{c|cll|ccc}
    \toprule
    Module & \multicolumn{3}{c|}{Method} & AP & $\text{AP}_{50}$ & $\text{AP}_{75}$ \\ \hline
    \multirow{2}{*}{BcS} & \multicolumn{3}{c|}{W/o adjustment} & 15.7 & 27.8 & \textbf{15.8} \\
   & \multicolumn{3}{c|}{W/ adjustment } & \textbf{15.9} & \textbf{28.2} & \textbf{15.8} \\ \hline
    \multirow{4}{*}{PG} & \multicolumn{3}{c|}{Cat-RPN} & 15.3 & 26.9 & 15.3 \\
     & \multicolumn{3}{c|}{Mult-RPN} & 15.3 & 27.2 & 15.3 \\
     & \multicolumn{3}{c|}{Cat-ROI} & \textbf{15.9} & 28.1 & \textbf{15.8} \\
     &\multicolumn{3}{c|}{Mult-ROI }& \textbf{15.9}& \textbf{28.2}& \textbf{15.8}\\ \bottomrule
    \end{tabular}
    \end{table}

\subsubsection{Impact of BcS module on the false prediction}
The BcS module is a core component of our approach, serving to eliminate false positives (FPs) and address the bias problem. To further analyze its impact, we report the number of FPs with respect to base classes both before and after the BcS module in Tab.\,\ref{tab4_process1}. Precision, formulated as $\frac{TP}{TP+FP}$, is also used to evaluate the performance in terms of suppressing FPs. Note that the experiments are performed on the split-2 of COCO dataset, which contains 3309 test images. The results indicate that the FPs are significantly reduced after the BcS module under different conditions and the precision is substantially improved, thus verifying the effectiveness of our BcS module.

\begin{figure}[!t]
  \centering
  {\includegraphics[width=3.2in]{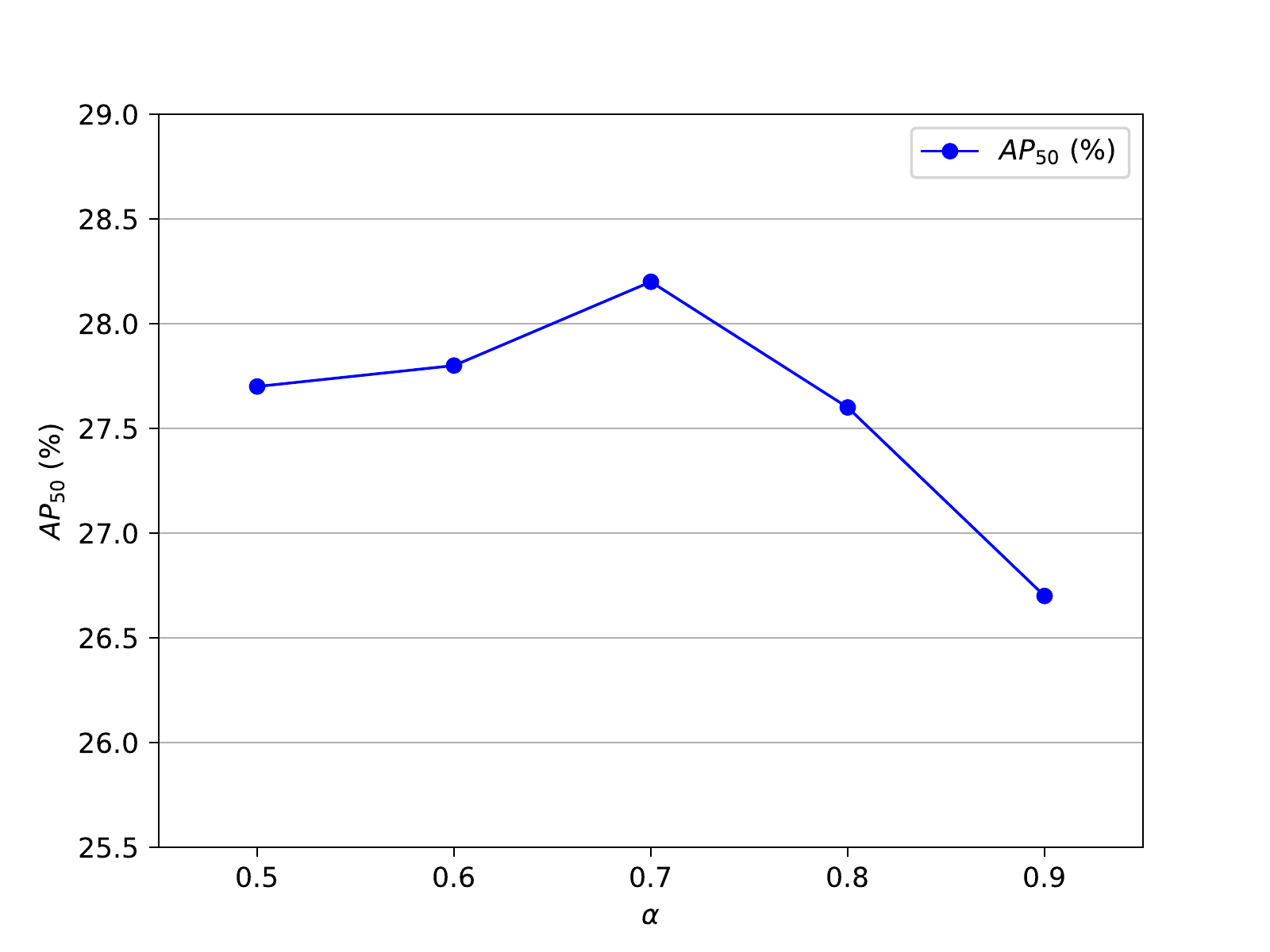}}
  \caption{Comparisons of different values of parameter $\alpha$ in the base-class suppression module on the split-2 of COCO dataset.}
  \label{fig:5}
\end{figure}

\begin{figure*}
  \centering
  {\includegraphics[width=6.5in]{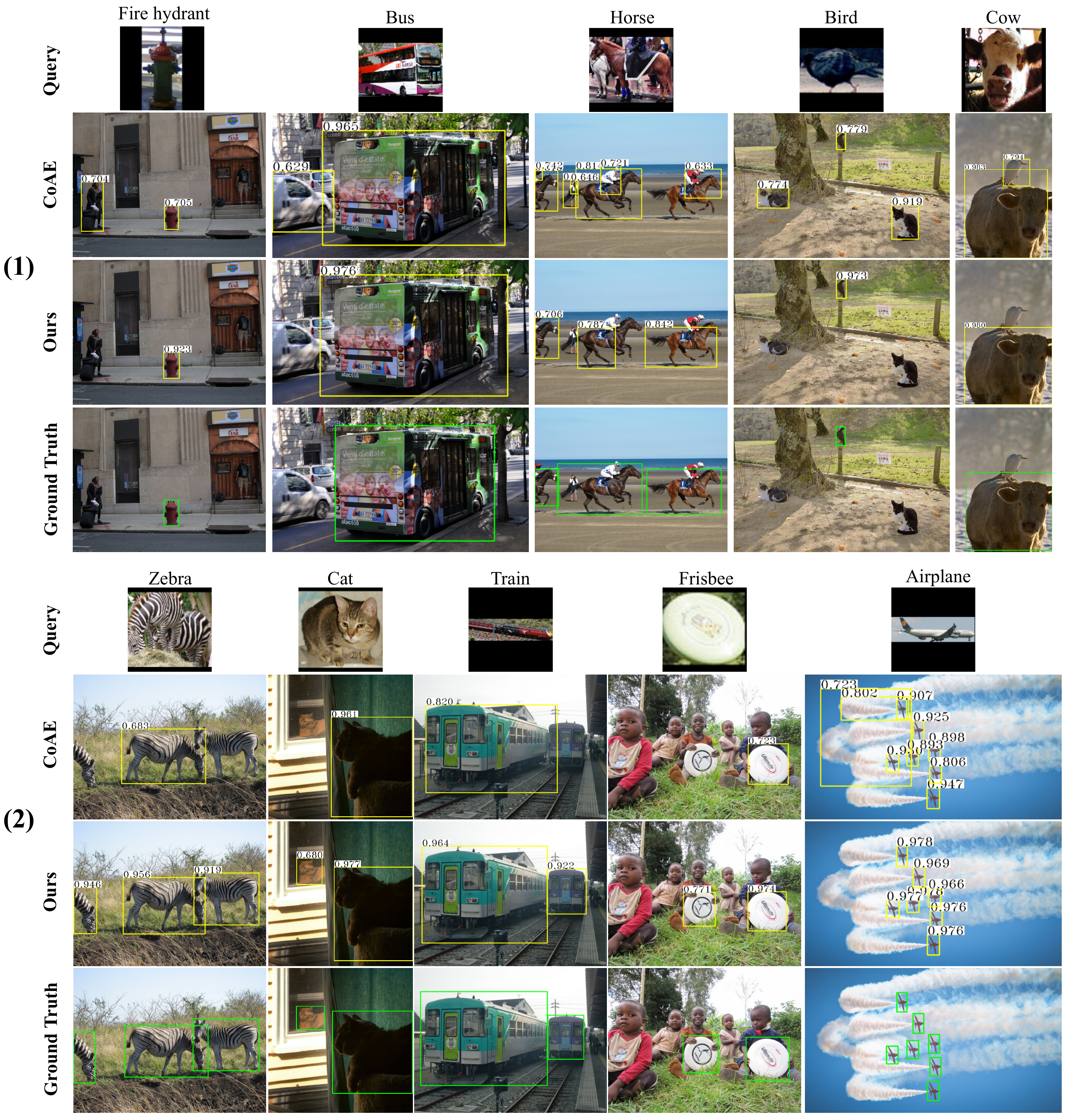}}
  \caption{Comparison of qualitative one-shot object detection results for novel classes on COCO dataset between CoAE\cite{NEURIPS2019_92af93f7} and our proposed approach. Each row from top to bottom denotes the query image, the results of CoAE and our model (yellow box), and ground truth (green box). In group (1), the target images include base-class distractors. The baseline method misclassifies the instances of base-classes as objects of interest, while our model can distinguish the target object from distractors. In group (2), the target images contain multiple objects that differ significantly in appearance, scale and shape from the specified query images. The baseline method produces false detection while our model presents relatively accurate locations. 
}
    \label{fig:4}
  \hfill
\end{figure*}

\subsubsection{Impact of each module}
Tab.\,\ref{tab4} shows the impact of the proposed Base-class Suppression (BcS) and Prior Guidance (PG) module on overall performance. Compared with the baseline, the BcS module brings a decent performance gain, proving its effectiveness in suppressing the false objects belonging to the base categories and mitigating the bias problem. The comparison of row 1 and row 3 shows that equipped with the PG module, the model benefits from the guidance of the prior map and yields 1.1$\%$ $\text{AP}_{50}$ gain over the baseline, which reveals its effectiveness. Integrating all modules, we achieve the best results and exceed the baseline by 2.7$\%$ $\text{AP}_{50}$, which demonstrates that our proposed modules indeed improve the ability to distinguish target objects from distractors. Specifically, the BcS module is applicable to complex scenarios consisting of diverse objects, while the PG module performs better in simple scenes containing salient objects. Thus, they complement each other to yield better results.

\subsubsection{Impact of base-class prediction adjustment}
To prevent false suppression on novel objects and achieve accurate suppression towards base classes, the results obtained from the base predictor are further adjusted under the guidance of simple similarity metrics. We evaluate the validity of the adjustment scheme and report the results in the first two rows of Tab.\,\ref{tab5}. It can be observed that the BcS module without adjustment leads to slightly inferior performance, indicating that the adjustment is an indispensable step to ensure effective base-class suppression. 

\subsubsection{Impact of different prior map guidance schemes}
The prior map plays a vital role in providing rich semantic cues and guiding the subsequent detection process. To explore the effect of different guidance strategies on the final detection performance, we respectively attempt to utilize the prior map to enrich the target features via concatenation or multiplication operation in the RPN or ROI head. Rows 3-6 of Tab.\,\ref{tab5} present the performance of our model under different guidance mechanisms, suggesting that performing prior guidance by multiplication in the ROI head is a better strategy.

\subsubsection{Impact of parameter $\alpha$ in the BcS module}
In order to precisely suppress the distractors belonging to base categories, we choose the appropriate proposals based on the classification score predicted by the base-class predictor. As mentioned in Section~\ref{3_3}, we select the proposals whose highest confidence scores over $S$ categories are more than threshold $\alpha$ as the base-class objects. We investigate the effect of the parameter $\alpha$ defined by Eq.\,(\ref{eq5}). Specifically, we vary $\alpha$ across a set of values $\left\{0.5,0.6,0.7,0.8,0.9\right\}$ to study its impact on the final performance on COCO dataset, as shown in Fig.\,\ref{fig:5}. Our model achieves the best results when the parameter $\alpha$ is set to 0.7. The appropriate choice of the threshold $\alpha$ can effectively mitigate the problem of model bias towards base classes. However, we also observe that larger values do not always yield more gains, and excessively large $\alpha$ can even lead to performance degradation. The reason could be that the base-class objects cannot be fully identified and further suppressed. 

\begin{table}[]
  \setlength\tabcolsep{12pt}
  \caption{Ablation study for different values of $\lambda_1$ and $\lambda_2$ in the loss function.}
  \label{tab6}        
  \renewcommand\arraystretch{1} 
  \centering  
  \begin{tabular}{ccll|ccc}
  \toprule
  $\lambda_1$ & \multicolumn{3}{c|}{$\lambda_2$} & AP & $\text{AP}_{50}$ & $\text{AP}_{75}$ \\ \hline
  0 & \multicolumn{3}{c|}{1} & 14.8  & 26.4 & 14.8 \\
  0.5 & \multicolumn{3}{c|}{0.5} & \textbf{15.9} & \textbf{28.2} & \textbf{15.8} \\
  1 & \multicolumn{3}{c|}{1} & 15.4 & 27.3 & 15.1 \\ \bottomrule
  \end{tabular}
  \end{table}

\subsubsection{Impact of parameter $\lambda_1$ and $\lambda_2$ in the loss function}
In the loss function defined by Eq.\,(\ref{eq8}), $\lambda_1$ and $\lambda_2$ are the weights assigned to $\mathcal{L}_{N}^{CE}$ and $\mathcal{L}_{F}^{CE}$, respectively. We investigate the impact of $\lambda_1$ and $\lambda_2$ on the final performance in Tab.\,\ref{tab6}. As can be observed in the first two rows, the results indicate that imposing supervision on the coarse results predicted by the novel predictor can facilitate further refinement and boost the final performance. Based on our experiments, we can conclude that when $\lambda_1$ and $\lambda_2$ are set to 0.5, our model yields the best performance.

%-------------------------------------------------------------------------
\subsection{Qualitative Results}
To better comprehend our proposed model, we visualize the detection results in Fig.\,\ref{fig:4}. Specifically, in group (1), the baseline method is easily distracted by base-class objects and misclassifies the distractors as objects of interest. In contrast, our model exhibits great superiority in distinguishing target objects from distractors, which is attributed to the base-class suppression module. Apart from the difficulties of the interference from the base-class objects, appearance variations and occlusions are also serious obstacles in the OSOD task. As shown in group (2), the baseline method neglects some objects in crowded scenes and produces false locations against complex backgrounds. While our model successfully identifies multiple objects and accurately localizes them despite significant variations in appearance, scale and shape between query and target images. These qualitative results clearly manifest that our proposed method can effectively tackle the challenge of base-class distractors, appearance variations and occlusions.

%------------------------------------------------------------------------
\section{Conclusion}
In this paper, targeting the phenomenon of model bias towards base classes and the generalization degradation on the novel classes, we rethink one-shot object detection from a new perspective and propose a novel BSPG network that combines a base-class predictor and a base-class suppression module to explicitly recognize and eliminate base-class objects. Furthermore, our PG module aims to leverage the class-agnostic prior map with rich semantic hints to guide the detection procedure and enhance overall performance. Extensive experiments demonstrate that our approach significantly improves the OSOD performance and achieves new state-of-the-art performance under various settings. We believe that our work offers valuable insights into addressing bias problems in the OSOD field and can inspire future research in this area.

\vfill
% \newpage
\bibliographystyle{IEEEtran}
\bibliography{mybib}
\end{document}